\documentclass[runningheads]{llncs}

\usepackage{eccv}

\usepackage{eccvabbrv}

\usepackage{multirow}
\usepackage{graphicx}
\usepackage{booktabs}
\usepackage{color}
\usepackage{colortbl}
\usepackage{pifont}
\newcommand{\cmark}{\ding{51}}%
\newcommand{\xmark}{\ding{55}}%

\usepackage{wrapfig}

\usepackage[accsupp]{axessibility}  

\usepackage[most]{tcolorbox}
\usepackage{inconsolata}
\usepackage[nosfdefault]{comicneue}
\usepackage{calligra}

\usepackage[pagebackref,breaklinks,colorlinks,citecolor=eccvblue]{hyperref}

\usepackage{orcidlink}

\begin{document}

\title{Holo-Captioning: \\Toward the Text Equivalent of 3D Scenes} 

\titlerunning{Holo-Captioning}


\author{Kun-Yu Lin\inst{1} \and
Chengke Bu\inst{1} \and
Zhenguo Li\inst{2} \and
Kai Han\inst{1}\thanks{Corresponding author.}}

\authorrunning{K.-Y.~Lin et al.}


\institute{Visual AI Lab, The University of Hong Kong \and
Frontier Robotics}

\maketitle

\begin{abstract}
This work introduces holo-captioning, a novel task that strives to seek the text equivalent of 3D scenes.
As the initial step, we formulate holo-captioning as generating a structured textual description that comprehensively depicts all entities within a 3D scene---including their semantic tags, spatial locations, attributes, and inter-entity relations. 
To tackle this challenging task, we first develop an effective captioning engine to produce detailed descriptions of individual entity instances and instance pairs, and contribute a large-scale benchmark comprising over 15K scenes for training and evaluation. 
Building upon this foundation, we propose HoloScribe, a novel model that features an instance-aware decoupled pipeline for generating structured holo-captions, and further incorporates anchor-aware instance linking to identify relational instance pairs. 
Additionally, we propose a comprehensive evaluation metric named HoloScore, and provide a human-curated test set to ensure reliable model assessment.
Experimental results demonstrate that HoloScribe significantly outperforms state-of-the-art 3D dense captioners and 3D LLM generalists, underscoring the effectiveness of our approach. 
Project page: \url{https://visual-ai.github.io/holocap/}
\end{abstract}

\section{Introduction}
\label{sec:intro}
\vskip -0.05in

\begin{figure}[t]
\centering
\includegraphics[width=1.0\linewidth]{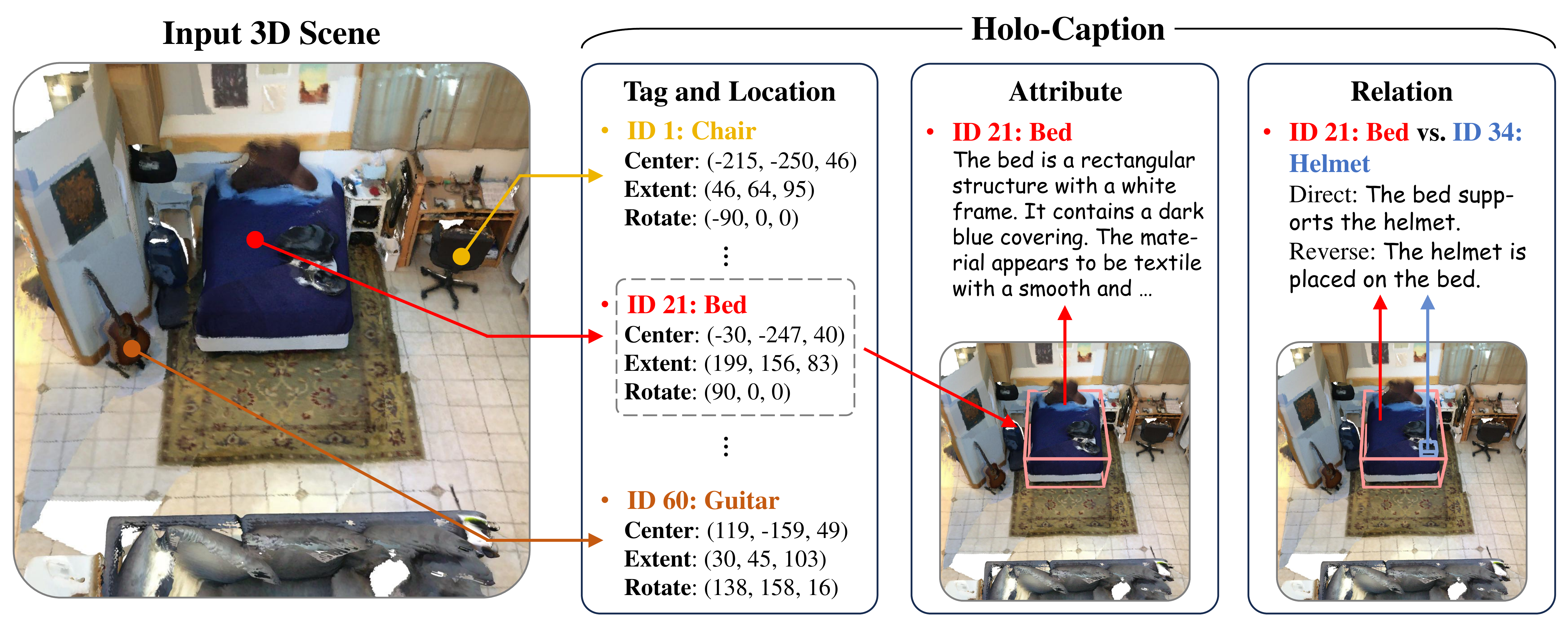}
\vskip -0.1in
\caption{
An example to demonstrate our proposed \textit{holo-captioning} task, which is formulated as generating a \textit{comprehensive structured textual description} for a 3D scene. 
A holo-caption comprehensively depicts \textit{all} entity instances within a 3D scene, including their respective \textit{semantic tags}, \textit{spatial locations} and \textit{attributes}, as well as \textit{relations} between entity instances. 
Best viewed in color.
}
\label{fig:teaser}
\vskip -0.15in
\end{figure}

Describing the 3D world with natural language is a fundamental field in computer vision and natural language processing, benefiting diverse 3D applications such as scene modeling~\cite{xie2025humaninthelooplocalcorrections3d,fang2025spatialgenlayoutguided3dindoor}, vision navigation~\cite{DBLP:journals/nca/WuCLYH24,DBLP:conf/rss/WerbyHBVB24} and robotic manipulation~\cite{DBLP:journals/corr/abs-2510-10903,DBLP:journals/corr/abs-2410-24164,DBLP:conf/cvpr/ZhouLLZ21,zhou2025explore}. 
As a pivotal task in this field, 3D dense captioning~\cite{Chen2021Scan2Cap,Chen2023Vote2Cap} aims to simultaneously detect and describe object instances within a 3D scene.
However, existing 3D dense captioning methods are constrained by a limited range of object categories and typically produce short, coarse textual descriptions. 
Driven by the rapid advances in Large Language Models (LLMs), some recent studies~\cite{10.1007/978-3-031-73030-6_14,mao2025spatiallmtraininglargelanguage} have proposed generating structured linguistic sequences to describe 3D scenes, thereby covering a broader range of categories in an open-vocabulary paradigm. 
Nevertheless, they primarily focus on detecting architectural layouts and salient objects, overlooking fine-grained entity attributes and inter-entity relations.
These omissions make existing methods inadequate for comprehensive scene understanding and limit their downstream applicability.
Motivated by these limitations, this work conceives of finding the \textit{text equivalent of a 3D scene}, an \textit{ambitious yet very challenging} goal, which aims to develop a \textit{comprehensive} textual description that captures all essential elements within a 3D scene, thereby offering great practical value and potentially facilitating diverse downstream applications.

In this work, we introduce holo-captioning, a novel task that strives to seek the text equivalent of 3D scenes. 
Our work serves as the \textit{initial step} toward this challenging objective. 
To render the problem tractable, we formulate holo-captioning as generating a structured textual description of a 3D scene, which \textit{comprehensively depicts all entity instances along with their semantic tags, spatial locations, attributes, and inter-entity relations}, as shown in Fig.~\ref{fig:teaser}. 
Compared with previous 3D dense captioners~\cite{Chen2021Scan2Cap,Chen2023Vote2Cap} and structural description generators~\cite{10.1007/978-3-031-73030-6_14,mao2025spatiallmtraininglargelanguage}, our formulation serves as a significantly closer approximation to the conceptual ``text equivalence'' due to its strong comprehensiveness.
Specifically, it demands not only precise entity localization but also fine-grained descriptions for entity attributes and inter-entity relations. 
More importantly, our formulation requires models to directly predict entity locations \textit{purely in textual form}, without relying on any auxiliary detectors or segmenters, thereby establishing a more intrinsic alignment between 3D scenes and text.

To systematically study holo-captioning, we propose HoloEngine, an effective captioning engine, and use it to construct HoloScan, a large-scale benchmark spanning over 15K indoor scenes across diverse categories.
Specifically, HoloEngine produces high-quality captions in an instance-centric manner. 
First, given 3D oriented bounding boxes from a 3D scene, HoloEngine projects these boxes into each camera view via the known intrinsics and extrinsics, yielding view-specific 2D boxes for each entity instance. 
Leveraging these view-specific boxes, we craft entity-aware prompts and instruct MLLMs to produce detailed view-specific descriptions for entity instances in perspective images. 
Finally, we use LLMs to perform viewpoint-aware consolidation, aggregating the view-specific texts into comprehensive captions.
Built with HoloEngine, the HoloScan benchmark features comprehensive structured holo-captions and is accompanied by a test set rigorously curated by human experts to ensure reliability.

To support comprehensive evaluation of holo-captioning, we propose HoloScore, a tailored metric that assesses model performance across four dimensions. 
Traditional captioning metrics, typically based on n-gram or lexical similarity (\eg, BLEU~\cite{DBLP:conf/acl/PapineniRWZ02}, CIDEr~\cite{DBLP:conf/cvpr/VedantamZP15}), fall short in evaluating holo-captions due to their substantial length and high degree of granularity. 
To address this, HoloScore employs an LLM-empowered procedure, leveraging the structured nature of holo-captions.
First, HoloScore performs grounded instance matching to assess semantic tagging and spatial localization. 
It then applies granular descriptor extraction to break long captions into atomic units. 
Finally, it evaluates the attribute and relation dimensions via dual descriptor matching.

Furthermore, we propose HoloScribe, a novel model to improve holo-captioning via an instance-aware decoupled pipeline.
Our model is motivated by the fact that holo-captions are long and information-dense, making it difficult for models to perceive all elements within 3D scenes in a single forward pass.
Accordingly, HoloScribe groups scene elements by entity instance and generates a structured textual description of the entire scene element by element. 
Specifically, HoloScribe first localizes entity instances and assigns semantic tags, and then generates detailed attribute and relation descriptions conditioned on the identified instances. 
To improve relation modeling, we further propose anchor-aware instance linking, which identifies relational instance pairs within a sampled subgraph conditioned on a given anchor instance. 
Our experiments demonstrate that HoloScribe significantly outperforms state-of-the-art 3D dense captioners and 3D LLM generalists in holo-captioning. 
Additionally, we evaluate HoloScribe in the 3D detection task on MultiScan~\cite{DBLP:conf/nips/MaoZJCS22}, a dataset unseen during training.
As shown in Table~\ref{tab:multiscan}, it achieves superior performance over two representative 3D LLMs in a
\begin{wraptable}{r}{0.32\textwidth} 
\vskip -0.4in
\fontsize{7.pt}{9.5pt}\selectfont
\setlength\tabcolsep{3.5pt}
\caption{
Tagging and localization F-scores on MultiScan. 
}
\label{tab:multiscan}
\centering
\begin{tabular}{c | c  c }
\hline 
Models
& Tag  & Loc   \\
\hline
LL3DA~\cite{Chen2024CVPR}
& 24.9  & 17.3     \\
~SpatialLM~\cite{mao2025spatiallmtraininglargelanguage}~
& 20.2  & 9.6      \\
\hline
\rowcolor{pink!20}
HoloScribe
& 53.5  & 20.2    \\
\hline
\end{tabular}
\vskip -0.35in
\end{wraptable}
zero-shot setting. 
Furthermore, we deploy HoloScribe for downstream mobile robotic object navigation tasks, where it achieves a 75\% success rate, surpassing the SpatialLM baseline (25\%). 
These results underscore the effectiveness and practical utility of our task and model.

In summary, our contributions are listed as follows:
(1) We introduce the novel task of holo-captioning, which aims to find the textual equivalent of 3D scenes. We take the initial step toward this ambitious goal by formulating it as generating comprehensive structured textual descriptions that cover four fundamental dimensions, and we contribute a reliable metric for evaluation. 
(2) We propose an effective instance-centric captioning engine to produce high-quality holo-captions, and contribute a large-scale benchmark of over 15K indoor scenes spanning diverse categories for training and evaluation. 
(3) We propose HoloScribe, a novel model that follows an instance-aware decoupled pipeline to generate comprehensive textual descriptions element by element. To the best of our knowledge, HoloScribe is the first 3D LLM capable of jointly localizing entity instances and generating detailed descriptions in \textit{pure text form} without extra detectors.

\section{Related Work}
\label{sec:related}
\vskip -0.05in

\textbf{3D Captioning.}
3D dense captioning~\cite{Chen2021Scan2Cap} is a foundational task in the 3D vision-language field that aims to simultaneously detect and describe object instances within a 3D scene.
As a pioneering work, Scan2Cap~\cite{Chen2021Scan2Cap} proposed an end-to-end method that leverages attention mechanisms and message-passing graphs to process point clouds and generate captions. 
In recent years, the field has witnessed substantial progress driven by a variety of technical innovations~\cite{Cai20223djcg,Chen2022D3Net,Chen2023Vote2Cap,DBLP:journals/pami/ChenZLCGLYLC24,Jiao2022MORE,Kim2024SeeItAll,Yuan2022XTrans2Cap}.
Despite these advances, existing models are built upon small-scale benchmarks (\eg, ScanRefer~\cite{DBLP:conf/eccv/ChenCN20}, Nr3D~\cite{DBLP:conf/eccv/AchlioptasAXEG20}), which restrict the range of object categories and yield short, coarse descriptions.
ExCap3D~\cite{excap2025} represents a notable step forward, proposing a fine-grained captioning benchmark and a model capable of generating both detailed object- and part-level descriptions.

Overall, previous works typically use benchmarks with fewer than 1K scenes and rely on 3D detectors or segmenters for object localization~\cite{DBLP:journals/inffus/GuoHJSSWHHSL26,Schult23ICRA,Chen2023Vote2Cap}.
\textit{In contrast, our work covers a substantially larger number of scenes and directly predicts all elements (including entity locations) in pure text form without requiring any additional models, marking a significant stride toward 3D-text alignment.}
Additionally, while holo-captioning is related to 3D object captioning, the latter focuses on generating descriptive sentences for isolated 3D objects~\cite{luo2023scalable,DBLP:conf/eccv/LuoJL24,ge2024visual,DBLP:conf/eccv/YanGYWXWZ24} rather than complex multi-object scenes, and heavily relies on 2D MLLMs with limited 3D spatial capabilities. 
Another related field is 3D scene graph (3DSG) generation, which represents a 3D scene as a graph consisting of object nodes and relation edges~\cite{DBLP:conf/cvpr/WaldDNT20,DBLP:conf/iccv/ArmeniHZGMFS19,DBLP:conf/cvpr/KochVCHR24,DBLP:conf/cvpr/Wu0C25}. 
3DSG fundamentally differs from our task, as it primarily focuses on predicting objects and relations, whereas capturing other aspects such as attributes typically requires auxiliary models.

More broadly, our work pursues the \textit{text equivalent of visual content}, a fundamental goal in vision-language research.
Our prior panoptic captioning work~\cite{lin2025panoptic} explores this goal for 2D images by representing an image as a purely textual description of entity instances, their locations and attributes, inter-entity relations, and the global image state.
Holo-captioning extends this pursuit to 3D scenes, a more challenging setting where visual-text alignment shifts from the 2D image plane to metric 3D space, requiring entities to be characterized by their 3D positions, spatial extents, orientations, attributes, and physical relations.
Together, panoptic captioning and holo-captioning are complementary steps toward our ultimate goal of visual-text equivalence.

\noindent\textbf{3D Large Language Models (3D LLMs).}
Inspired by advances in 2D Multi-modal LLMs (MLLMs)~\cite{DBLP:conf/nips/LiuLWL23a,DBLP:journals/corr/abs-2407-21783,DBLP:conf/aaai/WangSSLFTLTH025}, the field of 3D LLMs~\cite{DBLP:conf/ijcai/ZhaFYGC25} has witnessed remarkable progress in recent years. 
These models develop general 3D scene understanding capabilities through training on diverse 3D vision-language tasks, such as 3D question answering and 3D grounding~\cite{jia2024sceneverse,DBLP:conf/cvpr/WangMZXLLCZCXLL24,lyu2024mmscan,9878756,DBLP:conf/eccv/ChenCN20,DBLP:conf/eccv/AchlioptasAXEG20,Yang_2025_CVPR,huang2025surprise3ddatasetspatialunderstanding,ma2023sqad,linghu2024multimodal,huang20253dr1enhancingreasoning3d,zheng2025learning}. 
Existing 3D LLMs can be broadly categorized into two types based on their 3D vision representations. 
The first type relies on object-centric 3D representations~\cite{wang2023chat3ddataefficientlytuninglarge,DBLP:conf/iccv/ZhuMCD0023,huang2024chatscene,DBLP:conf/icml/HuangYMLLW0ZJ024}, \ie, these models typically first parse objects using 3D detectors or segmenters, then extract 3D object features, and use these object features as vision inputs.
The second type employs holistic scene-level representations without explicitly delineating objects~\cite{Chen2024CVPR,DBLP:conf/nips/HongZCZDCG23,DBLP:conf/eccv/ZhuZMNCJDHL24,Deng2025CVPR,DBLP:conf/cvpr/MeiLRWPW25}, \eg, LL3DA~\cite{Chen2024CVPR} develops a QFormer to connect 3D point clouds, visual prompts, and text. 
Additionally, motivated by the success of 2D MLLMs, some works attempt to construct 3D LLMs based on well-trained 2D video MLLMs, taking multi-view images as input~\cite{DBLP:conf/cvpr/ZhengH025,DBLP:journals/corr/abs-2501-01428,huang2025ThreeDRS}.
The development of 3D LLMs can benefit a wide range of downstream tasks, such as robotic tasks~\cite{DBLP:journals/nca/WuCLYH24,DBLP:conf/rss/WerbyHBVB24,DBLP:journals/corr/abs-2510-10903,DBLP:journals/corr/abs-2410-24164,wei2024grasp,wu2024economic,wei2025afforddexgrasp,DBLP:journals/corr/abs-2503-09186,wu2026vlanext,huang2026humanoid,li2024egoexo}.
Despite these advances, existing 3D LLMs are limited to handling relatively simple 3D vision-language tasks, \eg, generating concise captions and recognizing a restricted set of object categories. 
Pioneering efforts have explored pure textual representations for 3D scenes by developing specialized 3D LLMs~\cite{10.1007/978-3-031-73030-6_14,mao2025spatiallmtraininglargelanguage}. 
However, these works only focus on localizing and recognizing entity instances, while our holo-captioning aims to generate more comprehensive textual descriptions that incorporate attributes and relations.  
Another related work is Text-Scene~\cite{li2025textscenescenetolanguageparsingframework}, a hybrid approach that parses 3D scenes into textual descriptions using a suite of off-the-shelf models. 
This differs from our work, as we propose a single, unified model to produce comprehensive textual descriptions, with the goal of bridging 3D scenes and text.

\section{Holo-Captioning}
\label{sec:task}
\vskip -0.05in

\subsection{Task Formulation}
\label{subsec:task}

In this work, we formulate holo-captioning as generating a structured textual description that \textit{comprehensively} depicts all entity instances within a 3D scene, including their semantic tags, spatial locations, attributes, and the relations between entities. 
Unlike previous 3D captioning approaches~\cite{Chen2021Scan2Cap,Chen2023Vote2Cap}, our formulation encodes these four dimensions purely in text, resulting in a unified textual description that fully encapsulates the 3D scene.
In principle, holo-captioning considers four dimensions as follows:

\noindent\textit{\textbf{Semantic tag}} refers to the category label assigned to each entity instance in a 3D scene. 
We define ``entities'' as both architectural elements (\eg, floor, wall, window) and free-standing objects (\eg, table, bag, cabinet). 
Following SceneScript~\cite{10.1007/978-3-031-73030-6_14} and SpatialLM~\cite{mao2025spatiallmtraininglargelanguage}, we represent these category labels purely in text, without relying on class indices and predefined vocabularies. 

\noindent\textit{\textbf{Spatial location}} refers to the position and extent of an entity instance, represented by a 3D oriented bounding box. Specifically, in line with previous 3D object detection works~\cite{DBLP:conf/nips/DehghanBCFFGKDJ21,DBLP:conf/cvpr/Brazil0SR0G23,DBLP:conf/cvpr/WangMZXLLCZCXLL24}, we represent the oriented bounding box of an entity instance as a 9-DoF vector $(c_x, c_y, c_z, e_x, e_y, e_z, r_x, r_y, r_z)$, where $(c_x, c_y, c_z)$, $(e_x, e_y, e_z)$ and $(r_x, r_y, r_z)$ denote the center, size, and Euler-angle orientation of the box, respectively. 
By using oriented bounding boxes, a holo-caption can accurately describe the locations and spatial extents of instances in pure text, avoiding vague expressions such as ``at the center of the scene''. 
As detailed in Sec.~\ref{sec:method}, our work adopts a structured text form to specify semantic tags and spatial locations of instances. 
Notably, the direct prediction of precise spatial coordinates in a purely textual form, without the aid of auxiliary models, is highly non-trivial, necessitating strong 3D-text alignment capabilities.

\noindent\textit{\textbf{Attribute}} refers to characteristics or properties that describe the appearance or state of an entity instance. This dimension encompasses a wide range of attribute types, \eg, color, shape, material, texture, and constituent parts. Our work uses free-form text to describe attributes, which facilitates fine-grained instance identification and enhances the comprehensiveness of scene descriptions~\cite{DBLP:conf/eccv/ChenCN20,DBLP:conf/eccv/AchlioptasAXEG20,excap2025}. 

\noindent\textit{\textbf{Relation}} refers to the connections or interactions between two entity instances within a scene. This dimension encompasses a wide range of relation types, and our work focuses on those between nearby instances, as we primarily consider static indoor scenes where most meaningful relations occur among spatially adjacent entities. Typical examples include spatial relations (\eg, ``A is placed on B'') and state relations (\eg, ``A is leaning against B''). Similar to the attribute dimension, we describe inter-entity relations in free-form text, which facilitates a more comprehensive understanding of the structural organization and contextual semantics of 3D scenes~\cite{pmlr-v229-rana23a,10610243,DBLP:conf/corl/ChangBLCJKGAZBB23}. 

\noindent An example of our task formulation is presented in Fig.~\ref{fig:teaser}.
By considering these four fundamental dimensions, holo-captioning leads to a comprehensive textual description of a 3D scene.
Compared to previous works that pursue structural textual representations~\cite{10.1007/978-3-031-73030-6_14,mao2025spatiallmtraininglargelanguage}, holo-captions not only capture entity categories and locations, but also describe entity attributes and inter-entity relations, thereby enabling a comprehensive understanding of 3D scenes. 
Notably, holo-captioning requires a single model to directly predict all scene elements \textit{in pure text form} without relying on any additional models.
This significantly differs from previous 3D dense captioning works that require auxiliary detectors to predict entity locations~\cite{Chen2021Scan2Cap,Chen2023Vote2Cap,DBLP:journals/pami/ChenZLCGLYLC24,excap2025}. 
Such a formulation fosters a more intrinsic alignment between 3D scenes and text.

\subsection{The HoloScore Metric}
Previous 3D captioning works~\cite{Chen2021Scan2Cap,Chen2023Vote2Cap} primarily rely on traditional metrics like BLEU~\cite{DBLP:conf/acl/PapineniRWZ02}, METEOR~\cite{DBLP:conf/acl/BanerjeeL05}, and CIDEr~\cite{DBLP:conf/cvpr/VedantamZP15}. 
However, these metrics are tailored for short captions and have been shown to perform poorly on long, descriptive texts~\cite{ye2025painting,DBLP:journals/corr/abs-2405-19092}. 
To address this limitation, we propose HoloScore, a novel evaluation metric that comprehensively assesses holo-captions across four dimensions and achieves faithful alignment with our task objective.

\begin{figure}[t]
\centering
\includegraphics[width=1.0\linewidth]{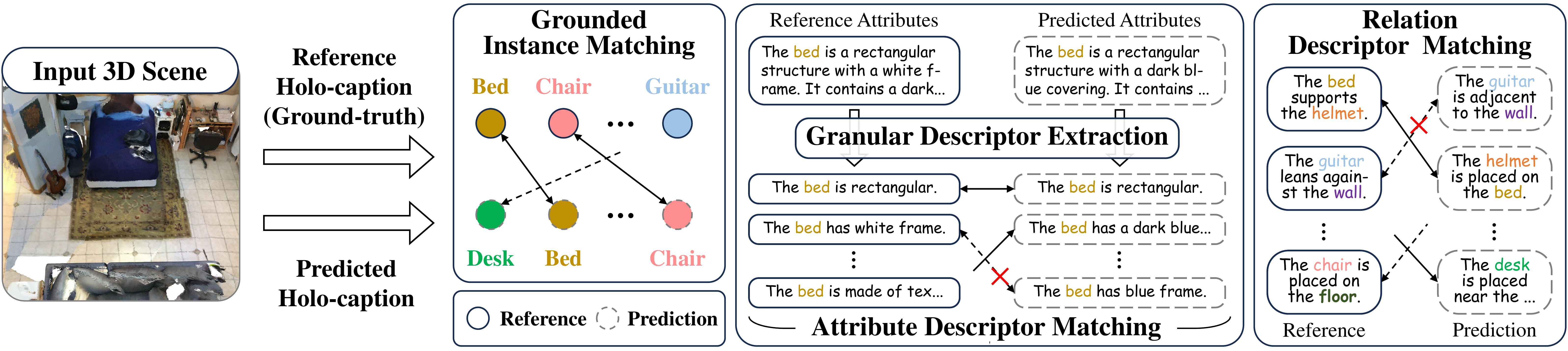}
\vskip -0.1in
\caption{
An overview of our HoloScore metric. 
HoloScore first groups holo-caption elements into three aspects: semantic tags with spatial locations, entity attributes and inter-entity relations. 
It then performs grounded instance matching to assess tagging and localization, applies granular descriptor extraction to process attributes, and performs dual descriptor matching to assess attribute and relation dimensions. 
}
\label{fig:metric}
\vskip -0.15in
\end{figure}

Given the structured nature of holo-captions, we group caption elements into three aspects, namely entity tags with spatial locations, entity attributes, and inter-entity relations. 
Based on this grouping, we design a three-step evaluation procedure comprising grounded instance matching, granular descriptor extraction, and dual descriptor matching. 
In what follows, we detail these three steps, and an overview of HoloScore is shown in Fig.~\ref{fig:metric}. 

\noindent \textbf{Grounded Instance Matching.} 
Given a set of ground-truth instances and a set of predicted ones, our grounded instance matching establishes a one-to-one correspondence between the two sets based on their semantic tags and spatial locations. 
Formally, let $\{(t_i, l_i)\}_{i=1}^n$ and $\{(\hat{t}_j, \hat{l}_j)\}_{j=1}^m$ denote the ground-truth and predicted entity instance sets, where $n$ and $m$ denote the total numbers of instances.
Here, $(t_i, l_i)$ represents the semantic tag and oriented bounding box of the $i$-th ground-truth instance, and $(\hat{t}_j, \hat{l}_j)$ represents those of the $j$-th prediction.  
Based on these definitions, we formulate grounded instance matching as an optimal bipartite matching problem:
\begin{equation*}
\begin{aligned}
    \Theta = \mathop{\arg\max}_{\Theta} \sum\nolimits_{i=1}^n \sum\nolimits_{j=1}^m (\alpha~\text{sim}_{i,j} + \mathrm{iou}(l_i, \hat{l}_j)) \cdot \Theta_{i,j}, \text{~s.t.~}\sum\nolimits_i\Theta_{i,j}\leq 1,~\sum\nolimits_j\Theta_{i,j}\leq 1, 
\end{aligned}
\end{equation*}
where $\Theta$ is a binary assignment matrix with $\Theta_{i,j}\in\{0,1\}$ indicating whether the $i$-th ground-truth instance is matched to the $j$-th prediction, \ie, $\Theta_{i,j}=1$ means a match.
$\text{sim}_{i,j}$ denotes the semantic similarity between $t_i$ and $\hat{t}_j$, computed using synonym matching and embedding similarity, while $\mathrm{iou}(l_i, \hat{l}_j)$ represents the 3D bounding-box IoU.
A weighting factor $\alpha=10$ is used to prioritize semantic tagging. 
In principle, two instances are considered a match when they share similar tags and are spatially close.

Based on the assignment matrix $\Theta$, we quantify model performance in semantic tagging and spatial localization. 
For semantic tagging, a matched pair is regarded as semantically consistent if their category tags share synonymous nouns or have similar text embeddings, \ie, $\text{sim}_{i,j} \geq \delta_\text{tag}$. 
By aggregating over all instances, we compute the precision and recall of semantic tagging and derive the corresponding F-score.
Precision measures the proportion of correctly tagged predictions, while recall measures the proportion of ground-truth entities correctly identified.
For spatial localization, two instances are regarded as location-consistent only if they are semantically consistent and their bounding boxes have an IoU above a preset threshold,
\ie, $\mathrm{iou}(l_i, \hat{l}_j) \geq \delta_\text{loc}$.
Following a similar protocol, we compute localization precision, recall, and the F-score to assess spatial localization quality.

\noindent \textbf{Granular Descriptor Extraction.}
To evaluate the nuanced attribute dimension more precisely, we propose a granular descriptor extraction method to decompose long and detailed textual descriptions into a set of atomic descriptors. 
Specifically, for the $i$-th ground-truth entity instance, our method takes its paired attribute description $A_i$ as input.  
We then prompt a state-of-the-art LLM to extract atomic descriptors from $A_i$, using carefully designed instruction prompts and in-context examples. 
This process is formulated as: $\{a_{i,k}\}_{k=1}^{n_i}=f_{\text{gde}}(A_i)$, where $a_{i,k}$ denotes the $k$-th atomic descriptor, $f_{\text{gde}}(\cdot)$ is the extraction function, and $n_i$ is the number of extracted descriptors of the $i$-th instance. 
Similarly, the extracted descriptors of the $j$-th predicted instance are denoted by $\{\hat{a}_{j,k}\}_{k=1}^{m_j}=f_{\text{gde}}(\hat{A}_j)$, where $\hat{A}_j$ denotes the predicted attribute description and $m_j$ is the number of extracted descriptors. 
This decomposition yields superior interpretability and controllability compared to directly evaluating long, unstructured descriptions.

\noindent \textbf{Dual Descriptor Matching.}
Finally, we evaluate model performance on the attribute and relation dimensions via descriptor matching. 
We employ an LLM-based embedding model~\cite{DBLP:journals/corr/abs-2506-05176} to encode text features, guided by a tailored prompt for robust similarity estimation between textual descriptors.
For the attribute dimension, we establish a one-to-one correspondence between the atomic descriptors of each ground-truth instance and its matched predicted instance.
The bipartite matching problem for the $i$-th ground-truth instance is formulated as follows:
\begin{equation*}
\begin{aligned}
    \Omega^i & = \mathop{\arg\max}_{\Omega^i} \sum\nolimits_{k=1}^{n_i} \sum\nolimits_{p=1}^{m_{\hat{i}}} \text{sim}^i_{k,p}  \cdot \Omega^i_{k,p}, \text{~s.t.~}\sum\nolimits_k\Omega^i_{k,p}\leq 1,~\sum\nolimits_p\Omega^i_{k,p}\leq 1,
\end{aligned}
\end{equation*}
where $\Omega^i$ is the assignment matrix of the $i$-th instance, and $\Omega^i_{k,p}=1$ indicates a match. 
$\text{sim}^i_{k,p}$ denotes the feature similarity between descriptors $a_{i,k}$ and $\hat{a}_{\hat{i},p}$, where the $\hat{i}$-th predicted instance is the matched one. 
A descriptor pair is considered consistent when $\text{sim}^i_{k,p}\geq\delta_{\text{attr}}$.
For each instance, we compute precision, recall, and the corresponding F-score, and then aggregate the F-scores across all instances to obtain the overall attribute metric.

For the relation dimension, we adopt a simplified procedure by direct descriptor matching, as relation descriptions are typically concise and each can be treated as a single atomic descriptor for efficiency. 
Given a ground-truth relation description $r_{i,j}$, we measure its similarity to the corresponding prediction $\hat{r}_{\hat{i},\hat{j}}$, where $\hat{i}$ and $\hat{j}$ are determined from the prior instance matching.
A relation pair is considered consistent if its similarity exceeds $\delta_{\text{rel}}$, and the overall F-score is derived in a similar manner to other dimensions.

In summary, our HoloScore metric comprises four F-scores, namely semantic tagging ($s_t$), spatial localization ($s_l$), entity attributes ($s_a$), and inter-entity relations ($s_r$). 
The final overall HoloScore is computed as $s = s_t + s_l + s_a + s_r$.

\section{Captioning Engine and Benchmark}
\label{sec:data}
\vskip -0.05in

\subsection{HoloEngine}
According to the task definition, we aim to generate comprehensive holo-captions that are both highly structured and instance-isolated. 
To this end, we develop HoloEngine, an instance-centric captioning engine that produces structured descriptions, facilitating seamless integration across instances. 
Overall, HoloEngine first localizes entity instances in multi-view images based on projected 2D bounding boxes, then generates view-specific attribute and relation descriptions using state-of-the-art Multi-modal Large Language Models (MLLMs), and finally aggregates view-specific descriptions into holistic captions via LLMs.

Specifically, for the attribute dimension, we first obtain 3D oriented bounding boxes from 3D scenes, and project them into 2D regions across multi-view images using known camera intrinsics and extrinsics.
Each 2D bounding box is overlaid on the image to explicitly indicate the target instance, and the category label is included in the text prompt (\eg, ``Describe the \underline{chair} within the highlighted box'').
This setup helps the MLLM focus on the correct instance and avoid confusion with nearby instances.
The MLLM then produces detailed attribute descriptions, guided by prompts that discourage horizontal spatial terms (\eg, ``left'', ``right'') and favor rotation-invariant phrasing (\eg, ``contain''), ensuring robustness to arbitrary scene orientations. 
Descriptions from all multi-view images are subsequently aggregated into a single, comprehensive caption by an LLM, conditioned on the corresponding viewing angles. 
To ensure efficiency, we sample entity-centric views at fixed angular intervals (\eg, every $30^{\circ}$) instead of using all possible viewpoints.

For the relation dimension, we generate captions for instance pairs similarly, with three key differences.
First, for each relational pair, we overlay a red and a blue bounding box in the image to explicitly indicate the subject and object instances, which we find significantly improves annotation quality in our preliminary validation.
Second, as most inter-entity relations occur between nearby instances, HoloEngine focuses exclusively on spatially proximate instance pairs, thereby improving efficiency.
Third, for each instance pair (\eg, A and B), we instruct the MLLM to produce relation descriptions from both perspectives, namely A relative to B (\eg, ``A is on top of B'') and B relative to A (\eg, ``B is below A''). 
This bidirectional strategy enriches the diversity of training data, and improves the comprehensiveness and reliability of evaluation.
Moreover, to ensure higher annotation quality, we apply an MLLM-based category refinement step before MLLM captioning on both dimensions, assigning more specific category labels to instances with coarse ones (\eg, replacing generic ``object'' labels).

\subsection{HoloScan}
Based on HoloEngine, we contribute a new benchmark named HoloScan for the holo-captioning task. 
We utilize publicly available 3D indoor scene scans from five data sources, including four real datasets (ScanNet~\cite{8099744}, 3RScan~\cite{9010673}, Matterport3D~\cite{8374622} and ARKitScenes~\cite{DBLP:conf/nips/DehghanBCFFGKDJ21}) and one synthetic dataset Structured3D~\cite{10.1007/978-3-030-58545-7_30}. 
This results in a large-scale benchmark comprising over \textit{8.4K real 3D scans} and 7.5K synthetic scenes. 
Its category labels and 3D bounding boxes are annotated with reference to EmbodiedScan~\cite{DBLP:conf/cvpr/WangMZXLLCZCXLL24}, MMScan~\cite{lyu2024mmscan} and Structured3D.
Following previous works~\cite{DBLP:conf/cvpr/WangMZXLLCZCXLL24,lyu2024mmscan,jia2024sceneverse}, we standardize the data 
format, scene scale, sampling frequency, and viewpoint configuration across all datasets.
In addition, we utilize only the spatial location information of Structured3D to improve training, as its synthetic instance appearance differs from that of real scenes. 
During data production, we select the best MLLM for each step through preliminary experiments, balancing image understanding, instruction-following, and inference efficiency.
Overall, HoloScan comprises over 13K training scenes and 619 validation scenes paired with high-quality auto-generated holo-captions, along with 83 
\begin{wraptable}{r}{0.6\textwidth} 
\vskip -0.37in
\fontsize{7.pt}{9.5pt}\selectfont
\setlength\tabcolsep{1pt}
\caption{
Comparison with 3D dense captioning and structural description generation benchmarks.
}
\label{tab:data}
\centering
\begin{tabular}{c | c | c c  c c }
\hline 
Data 
& Type  & ~~Box~~      & Category & ~Scene~ & ~Word~  \\
\hline
ScanRefer~\cite{DBLP:conf/eccv/ChenCN20}
& Real  & \cmark     & 18       & 0.7K  & 17.9   \\
Nr3D~\cite{DBLP:conf/eccv/AchlioptasAXEG20}
& Real  & \cmark     & 18       & 0.6K  & 11.4   \\
ExCap3D~\cite{excap2025}
& Real  & \xmark     & 84       & 0.9K  & 82.8   \\
ASE~\cite{10.1007/978-3-031-73030-6_14}
& Synthetic  & \cmark     & Layout   & 100K  & \xmark \\
SpatialLM~\cite{mao2025spatiallmtraininglargelanguage}
& Synthetic  & \cmark     & 62       & 54K   & \xmark \\
\hline
\rowcolor{pink!20}
HoloScan
& Hybrid  & \cmark     & \textbf{734}  & 15K  & \textbf{89.0} \\
\hline
\end{tabular}
\vskip -0.25in
\end{wraptable}
test scenes paired with holo-captions rigorously curated by human experts.

In summary, HoloScan features high-quality structured holo-captions that comprehensively depict scene elements along four fundamental dimensions. 
As shown in Table~\ref{tab:data}, compared with previous 3D dense captioning benchmarks~\cite{DBLP:conf/eccv/ChenCN20,DBLP:conf/eccv/AchlioptasAXEG20,excap2025}, HoloScan provides longer and more detailed instance descriptions (more words on average) and encompasses a broader range of entity categories. 
Moreover, it covers a wider variety of \textit{real scenes} and sources data from multiple datasets, unlike previous benchmarks that rely solely on a single source (\eg, ScanNet~\cite{8099744}).
Additionally, unlike previous structural description generation benchmarks that focus primarily on \textit{synthetic} scenes~\cite{10.1007/978-3-031-73030-6_14,mao2025spatiallmtraininglargelanguage}, HoloScan contains a substantial number of \textit{real-world scenes} and provides \textit{comprehensive descriptions} with fine-grained coverage of attributes and relations. 
Collectively, these advancements position our task formulation as a significantly closer approximation to the conceptual ``text equivalence''.

\section{HoloScribe}
\label{sec:method}
\vskip -0.05in

\begin{figure}[t]
\centering
\includegraphics[width=1.0\linewidth]{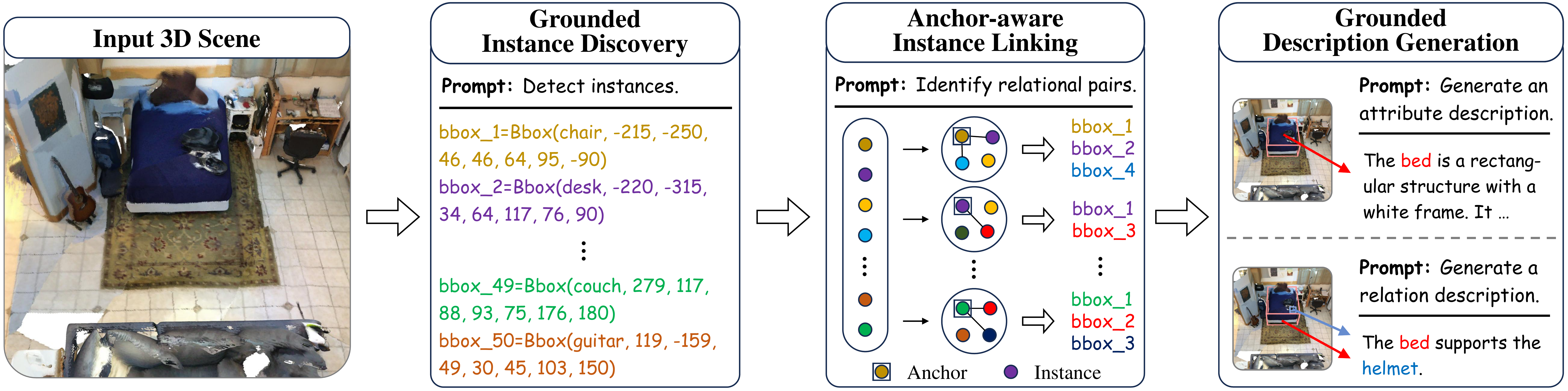}
\vskip -0.1in
\caption{
An overview of our HoloScribe. 
Given an input 3D scene, HoloScribe generates holo-captions following an instance-aware decoupling pipeline, consisting of grounded instance discovery, anchor-aware instance linking, and grounded description generation. 
}
\label{fig:method}
\vskip -0.1in
\end{figure}

To address holo-captioning, we propose HoloScribe, a novel model that follows an instance-aware decoupled pipeline. 
The design of HoloScribe is motivated by the fact that holo-captions are long and information-dense, making it difficult for models to perceive all elements in 3D scenes in a single forward pass. 
To tackle this challenge, HoloScribe decomposes the captioning process into instance-centric subtasks, progressively generating the structured textual description of a scene element by element.
Specifically, HoloScribe first localizes entity instances and assigns semantic tags, and then generates detailed attribute and relation descriptions conditioned on the identified instances. 
An overview of HoloScribe is shown in Fig.~\ref{fig:method}. 

Formally, HoloScribe takes as input a 3D scene in point cloud format, denoted by $\mathbf{Q}_v$, and produces a holo-caption $\hat{\mathbf{H}}$ via three phases, namely grounded instance discovery, anchor-aware instance linking, and grounded description generation, which we detail below. 
Following SpatialLM~\cite{mao2025spatiallmtraininglargelanguage}, HoloScribe adopts a LLaVA-style 3D LLM architecture, consisting of a point cloud encoder, an MLP connector, and a decoder-only LLM. 
In each phase, we employ a specific prompt to guide HoloScribe in producing desired outputs in pure text form.

\noindent \textbf{Grounded Instance Discovery.}
This phase aims to identify all entity instances within the input scene, serving as the foundation for a holo-caption.
For training, we first extract entity instances along with their semantic tags and spatial locations from the ground-truth holo-caption ${\mathbf{H}}$.
We then construct a point-text training tuple $(\mathbf{Q}_v,\mathbf{Q}_{\text{inst}},\mathbf{H}_{\text{inst}})$, where $\mathbf{Q}_{\text{inst}}$ denotes the tailored prompt and $\mathbf{H}_{\text{inst}}$ is a holo-instance text composed of individual instance texts. 
During inference, this phase outputs a predicted holo-instance text $\hat{\mathbf{H}}_{\text{inst}}$ given a scene.
Each instance in $\mathbf{H}_{\text{inst}}$ is represented by a structured text of the format
``\texttt{bbox\_$i$=Bbox($t_i$,$c_{i,x}$,$c_{i,y}$,$c_{i,z}$,$e_{i,x}$,$e_{i,y}$,$e_{i,z}$,$r_{i,z}$)}'',
where $i$ denotes the instance index, $t_i$ is the semantic tag, and the remaining terms are box parameters as defined in Sec.~\ref{subsec:task}.
To simplify learning, we adopt 7-DoF bounding boxes during training, which are recomputed from each entity's point cloud by constraining $r_{i,x}=r_{i,y}=0$.
Given that real-world scenes often contain dozens of instances, we further divide them into three types, namely \textit{architectural elements}, \textit{common objects}, and \textit{other objects}. 
During inference, we predict each type independently with tailored prompts and then aggregate the results.

\noindent \textbf{Anchor-aware Instance Linking.}
Given the detected entity instances, this phase aims to identify instance pairs that exhibit inter-entity relations.
A straightforward approach is to predict relations for every instance pair in the scene.
However, since typical scenes often contain dozens of instances, exhaustively considering every possible pair would result in an excessive number of relation candidates and high computational cost. 
To address this, we propose an anchor-aware edge formation strategy that learns relation modeling within sampled subgraphs of the complete instance graph.
This design enables HoloScribe to efficiently focus on instance pairs that are likely to exhibit meaningful relationships, and to generate relation descriptions only for these selected pairs.

Given a scene, we first construct a graph where each entity instance corresponds to a node.
The goal of anchor-aware instance linking is to determine whether an edge should be formed between two nodes, indicating the presence of a relationship between the corresponding instances.
Since the complete instance graph can be large, we instead sample smaller subgraphs and perform instance linking within these subgraphs to make relation learning more tractable.
During training, we sample a large and diverse set of subgraphs to improve learning coverage. 
This yields a set of point-text tuples $\{(\mathbf{Q}_v,\mathbf{Q}_\text{pair},\mathbf{H}_{\text{subg}},\mathbf{H}_{\text{pair}})\}$ from the ground-truth instance set,
where $\mathbf{Q}_\text{pair}$ is the prompt, $\mathbf{H}_\text{subg}$ is text describing a sampled subgraph, and $\mathbf{H}_\text{pair}$ is a relation text specifying relationship existence. 

For each subgraph, we designate one instance as the anchor and sample $n_g-1$ additional instances to form a subgraph with $n_g$ nodes in total. 
The subgraph is then serialized into $\mathbf{H}_\text{subg}$ and fed into HoloScribe. 
Its textual format mirrors the holo-instance text used in grounded instance discovery, listing the $n_g$ sampled instances with the anchor instance always placed first. 
The corresponding target text $\mathbf{H}_\text{pair}$ specifies which of the remaining instances are related to the anchor.
For example, the relation text ``\texttt{bbox\_1,bbox\_2,bbox\_3}'' indicates that the second and third instances are related to the anchor (\ie, the first instance).
To ensure decoding stability and prevent empty outputs, we enforce that each relation text always begins with ``\texttt{bbox\_1}''.
During inference, each discovered instance is treated as the anchor in turn, and we construct subgraphs $\{\hat{\mathbf{H}}_\text{subg}\}$ accordingly. 
By aggregating the predicted relation texts $\{\hat{\mathbf{H}}_\text{pair}\}$ across all anchor-centric subgraphs, we obtain the complete set of relational instance pairs for the entire scene.

\begin{table}[t]
\fontsize{7.5pt}{10.5pt}\selectfont
\setlength\tabcolsep{3.0pt}
\caption{
Comparison results on the validation and test sets of HoloScan. 
Performance is measured by HoloScore, and we report the scores in tagging, location, attribute, relation, and the overall score. 
\textbf{Bold} indicates the best, and ``-'' means the model does not support that dimension. 
}
\vskip -0.1in
\label{tab:sota}
\centering
\begin{tabular}{c | c | c  c  c c | c }
\hline
Models & LLM & Tagging & Location & Attribute & Relation & Overall \\
\hline
\rowcolor{gray!6}
\multicolumn{7}{l}{\textbf{~~~~~~Validation Set}} \\
\hline
Vote2Cap-DETR~\cite{Chen2023Vote2Cap} & -
& 32.87 & 17.00 & 2.49  & -     & -     \\
~Vote2Cap-DETR++~\cite{DBLP:journals/pami/ChenZLCGLYLC24}~ & -
& 30.84 & 16.70 & 2.38  & -     & -     \\
LEO~\cite{DBLP:conf/icml/HuangYMLLW0ZJ024} & Vicuna-7B
& 23.34 & 14.25 & 1.53  & -     & -     \\
LL3DA~\cite{Chen2024CVPR} & OPT-1.3B
& 31.28 & 16.50 & 2.99  & -     & -     \\
SpatialLM-Tuned~\cite{DBLP:conf/icml/HuangYMLLW0ZJ024} & Qwen2.5-0.5B
& 51.73 & 9.94  & 6.74  & 2.60 & 71.01 \\
LL3DA-Tuned~\cite{Chen2024CVPR} & Qwen2.5-0.5B
& 31.25 & 16.50 & 10.36 & 4.83 & 62.94 \\
\rowcolor{pink!20}
HoloScribe (Ours) & ~Qwen2.5-0.5B~
& \textbf{60.43} & \textbf{22.33} & \textbf{22.52} & \textbf{11.08} & \textbf{116.36} \\
\hline
\rowcolor{gray!6}
\multicolumn{7}{l}{\textbf{~~~~~~~~~Test Set}} \\
\hline
Vote2Cap-DETR~\cite{Chen2023Vote2Cap} & -
& 32.50 & 17.71 & 2.70  & -     & -     \\
~Vote2Cap-DETR++~\cite{DBLP:journals/pami/ChenZLCGLYLC24}~ & -
& 31.39 & 17.04 & 2.67  & -     & -     \\
LEO~\cite{DBLP:conf/icml/HuangYMLLW0ZJ024} & Vicuna-7B
& 25.91 & 14.93 & 1.60  & -     & -     \\
LL3DA~\cite{Chen2024CVPR} & OPT-1.3B
& 31.28 & 16.89 & 3.27  & -     & -     \\
SpatialLM-Tuned~\cite{DBLP:conf/icml/HuangYMLLW0ZJ024} & Qwen2.5-0.5B
& 49.82 & 9.23  & 7.01  & 3.57 & 69.63 \\
LL3DA-Tuned~\cite{Chen2024CVPR} & Qwen2.5-0.5B
& 31.28 & 16.88 & 9.54  & 3.97 & 61.67 \\
\rowcolor{pink!20}
HoloScribe (Ours) & ~Qwen2.5-0.5B~
& \textbf{61.83} & \textbf{22.75} & \textbf{23.72} & \textbf{8.74} & \textbf{117.04} \\
\hline
\end{tabular}
\end{table}

\noindent \textbf{Grounded Description Generation.}
After discovering entity instances and identifying relational pairs, we proceed to generate detailed attribute and relation descriptions based on entity tags and locations. 
For the attribute dimension, we construct point-text tuples $\{(\mathbf{Q}_v,\mathbf{Q}_\text{attr},\mathbf{H}_i,\mathbf{H}^a_i)\}$ for training, where $\mathbf{Q}_\text{attr}$ is the prompt. 
$\mathbf{H}_i$ is a text indicating the semantic tag and spatial location of the $i$-th ground-truth instance, resembling holo-instance texts. 
$\mathbf{H}^a_{i}$ is the corresponding ground-truth attribute description, which serves as the target output of the attribute dimension. 
Similarly, for the relation dimension, we construct point-text tuples $\{(\mathbf{Q}_v,\mathbf{Q}_\text{rel},\mathbf{H}_{i,j},\mathbf{H}^r_{i,j})\}$ for training.
$\mathbf{Q}_\text{rel}$ is the prompt, $\mathbf{H}_{i,j}$ is the instance text composed of the $i$-th and $j$-th instances, and $\mathbf{H}^r_{i,j}$ is the corresponding ground-truth relation description. 
During inference, we take the discovered instances $\{\hat{\mathbf{H}}_{i}\}$ and identified instance pairs $\{\hat{\mathbf{H}}_{i,j}\}$ as inputs, and generate attribute descriptions $\{\hat{\mathbf{H}}^a_{i}\}$ and relation descriptions $\{\hat{\mathbf{H}}^r_{i,j}\}$. 
Finally, by aggregating model outputs across all four dimensions, HoloScribe produces complete holo-captions in a structured text form.

\section{Experiment}
\label{sec:experiment}
\vskip -0.05in

\noindent \textbf{Implementation Details.} 
Following SpatialLM~\cite{mao2025spatiallmtraininglargelanguage}, our HoloScribe adopts an ``Encoder-MLP-LLM'' architecture with Sonata~\cite{DBLP:conf/cvpr/0002DFSXYENZS25} as the vision encoder, and it is initialized using the SpatialLM1.1-Qwen-0.5B checkpoint. 
We finetune our model on the training set of HoloScan for one epoch using LoRA. 
For HoloEngine, we use Gemma3-12B~\cite{gemmateam2025gemma3technicalreport} to produce descriptions of images and Qwen3-4B-Thinking~\cite{yang2025qwen3technicalreport} to aggregate descriptions. 
For HoloScore, we use Gemma3-1B~\cite{gemmateam2025gemma3technicalreport} for granular descriptor extraction, and use Qwen3-Embedding-0.6B~\cite{DBLP:journals/corr/abs-2506-05176} for encoding text in dual descriptor matching. 
The coefficients are set as $\delta_{\text{tag}}=0.5$, $\delta_\text{loc}=0.3$, and $\delta_\text{attr}=\delta_\text{rel}=0.6$. 

\noindent \textbf{Comparison with State-of-the-Arts.} 
Table~\ref{tab:sota} summarizes the comparison with state-of-the-art captioning models, including 3D dense captioners~\cite{Chen2023Vote2Cap,DBLP:journals/pami/ChenZLCGLYLC24}, 3D LLM generalists~\cite{DBLP:conf/icml/HuangYMLLW0ZJ024,Chen2024CVPR}, and holo-captioning specialists (see the Appendix for baseline details). 
From the table, we find that both 3D dense captioners and 3D LLM generalists obtain poor performance in holo-captioning, especially on the attribute dimension. 
This is because these models are trained on traditional captioning benchmarks (\eg, ScanRefer~\cite{DBLP:conf/eccv/ChenCN20}) and thus can only produce short captions describing simple attributes like color or shape. 
Compared with holo-captioning specialists trained on HoloScan, our HoloScribe obtains significant improvements on all dimensions, which is attributed to our proposed instance-aware decoupled pipeline.

\begin{table}[t]
\fontsize{7.5pt}{10.5pt}\selectfont
\setlength\tabcolsep{3.0pt}
\caption{
Ablation study of our HoloScribe on the validation set of HoloScan. 
}
\vskip -0.1in
\label{tab:ablation}
\centering
\begin{tabular}{c | c c c c | c }
\hline
Models & ~Tagging~ & Location & Attribute & ~Relation~ & ~Overall~ \\
\hline
Baseline  
& 46.98 & 8.23  & 2.39  & 1.04 & 58.64 \\
w/ decoupling
& 51.73 & 9.94  & 6.74  & 2.60 & 71.01 \\
w/ instance-aware prediction
& 59.20 & 20.45 & 22.41 & 7.92 & 109.98 \\
Full w/o anchor
& 58.67 & 17.94 & 22.46 & 9.20 & 108.27 \\
\rowcolor{pink!20}
Full
& \textbf{60.43} & \textbf{22.33} & \textbf{22.52} & \textbf{11.08} & \textbf{116.36} \\
\hline
\end{tabular}
\end{table}

\noindent \textbf{Ablation Study.} 
Table~\ref{tab:ablation} summarizes the ablation study of HoloScribe. 
We adopt the directly tuned SpatialLM as the baseline, which performs poorly because the desired holo-captioning outputs are long and highly content-rich.
After isolating the relation prediction component, we observe a clear performance improvement, as shown by ``w/ decoupling''.
By introducing the instance-aware attribute and relation prediction mechanism (``w/ instance-aware prediction''), the overall performance of HoloScribe improves significantly, where relation prediction is performed for every instance pair.
By further incorporating our anchor-aware linking mechanism (``Full''), HoloScribe obtains substantial improvement on the relation dimension, as the model can identify meaningful relational pairs while filtering out irrelevant pairs, thereby reducing erroneous predictions. 
Finally, we conduct an additional comparison against a variant without the anchor mechanism (``Full w/o anchor''), \ie, one that predicts all relational pairs within each subgraph.
The results demonstrate the effectiveness of our anchor-aware design, as the anchor-free variant struggles to capture relations among all possible pairs simultaneously.

\begin{figure}[t]
\centering
\includegraphics[width=0.95\linewidth]{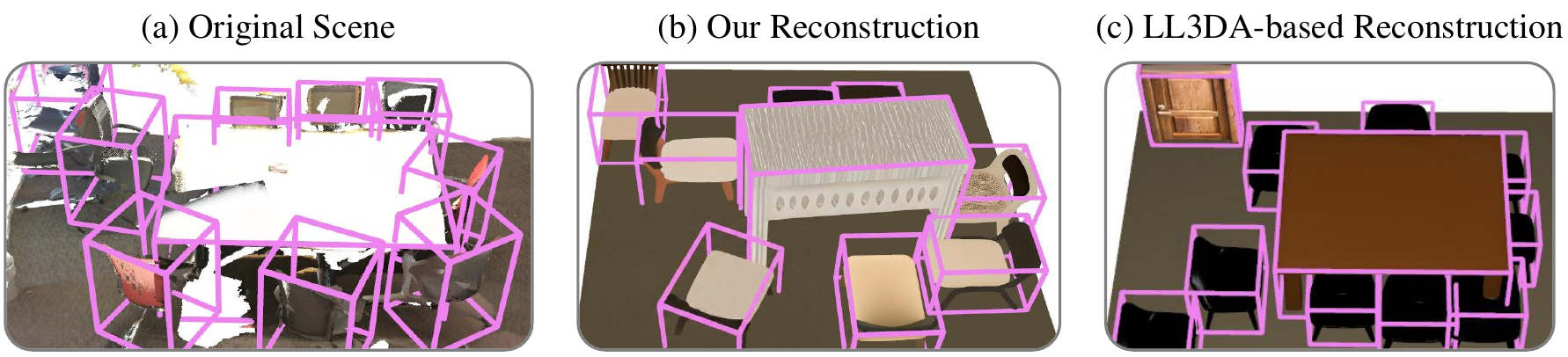}
\vskip -0.1in
\caption{
Demonstration of (a) a real scene from HoloScan, (b) the HoloScribe-based reconstruction, and (c) LL3DA-based reconstruction. 
We highlight entities by 3D boxes.
}
\label{fig:reconstruction}
\vskip -0.1in
\end{figure}

\noindent \textbf{3D Scene Reconstruction.} 
We conduct a scene reconstruction experiment to qualitatively demonstrate the effectiveness and value of our model. 
Specifically, we employ Hunyuan3D~\cite{DBLP:journals/corr/abs-2506-15442} to generate 3D objects using instance tags and attributes derived from a generated holo-caption.
The size and pose of each object are then refined according to the predicted bounding boxes, and the objects are positioned in 3D space to assemble the reconstructed scene.
As shown in Fig.~\ref{fig:reconstruction} (a) and (b), HoloScribe produces reconstructed 3D scenes that closely match the original ones, particularly in terms of accurate semantic tagging and precise entity localization. 
As shown in Fig.~\ref{fig:reconstruction} (c), our reconstruction outperforms the LL3DA-based one, as LL3DA suffers from inaccurate entity detection (yielding redundant, incorrect chairs) and monotonous attributes (usually predicting the single ``black'' attribute).

\noindent \textbf{Cross-Dataset Detection.}
To demonstrate the generalizability of HoloScribe, we conduct a detection experiment using MultiScan~\cite{DBLP:conf/nips/MaoZJCS22}, which is unseen during training. 
Specifically, for each scene, we apply HoloScribe to produce a holo-caption, from which we extract the entity tags and locations to perform the detection task. 
As shown in Table~\ref{tab:multiscan}, two representative 3D LLMs, namely LL3DA and SpatialLM, perform poorly on MultiScan.  
This is because LL3DA can only detect a limited range of categories, and SpatialLM is trained solely on synthetic scenes and thus hardly generalizes to real scenes.
In contrast, HoloScribe significantly outperforms these baselines as it can recognize diverse categories and generalize more effectively to real scenes from unseen datasets. 
Additional results in the Appendix show that holo-captions enable flexible scene editing.

\section{Conclusion}
\label{sec:conclusion}
\vskip -0.05in

This work introduces holo-captioning, a novel task that pursues the text equivalent of 3D scenes. 
We take the initial step toward this ambitious goal by formulating it as the generation of a structured textual description that comprehensively depicts scene elements across four dimensions. 
To enable systematic study, we propose a comprehensive evaluation metric, develop an effective captioning engine, and contribute a large-scale benchmark. 
We further propose HoloScribe, which follows an instance-aware decoupled pipeline to address holo-captioning. 
To our knowledge, HoloScribe is the first 3D LLM capable of jointly localizing entity instances and generating detailed descriptions purely in text, demonstrating strong performance in extensive experiments. 

Despite our efforts, this work represents only the initial attempt toward our conceptual goal of ``text equivalence'', and holo-captioning still presents numerous open challenges. 
Our experiments reveal that the task remains highly challenging for existing models, partially due to the inherent difficulty of predicting entity locations in pure text form. 
We consider scaling up the model a promising direction for future work.
Nevertheless, this work establishes a solid foundation for future research and is expected to catalyze further advances in bridging 3D scenes and text.

\noindent\textbf{Acknowledgments.}
This work is supported by the Hong Kong Research Grants Council - General Research Fund (Grant No.: $17211024$), the Innovation and Technology Fund (ITS/488/24FP), and the HKU Seed Fund for PI Research.
The authors would like to thank Yi-Xiang He and Yi-Lin Wei for their support in downstream experiments. 
The authors also thank Jiaming Zhou, Yu-Ming Tang, and Weining Ren for their valuable suggestions on the writing. 
The authors also thank Yuxian Li, Mingshan Huang, Huairong Chen, Shenru Zhang, Yixuan Chen, Junming Chen, Yifei Zhang, Zhenyi Fan, Jialu Tang, and Boning Shao for their help and support for our project.

\appendix

\section*{Appendix}

\renewcommand{\thesection}{A\arabic{section}}
\setcounter{equation}{0}    

\renewcommand{\thefigure}{A\arabic{figure}} 
\setcounter{figure}{0} 
\renewcommand{\thetable}{A\arabic{table}}   
\setcounter{table}{0}  
\renewcommand{\theequation}{A\arabic{equation}}
\setcounter{equation}{0}


\section{More Details of Task Formulation}
\label{sec:taskexample}
\vskip -0.05in

\subsection{Annotation Examples}
\begin{figure}[h]
\vskip -0.1in
\centering
\includegraphics[width=1\linewidth]{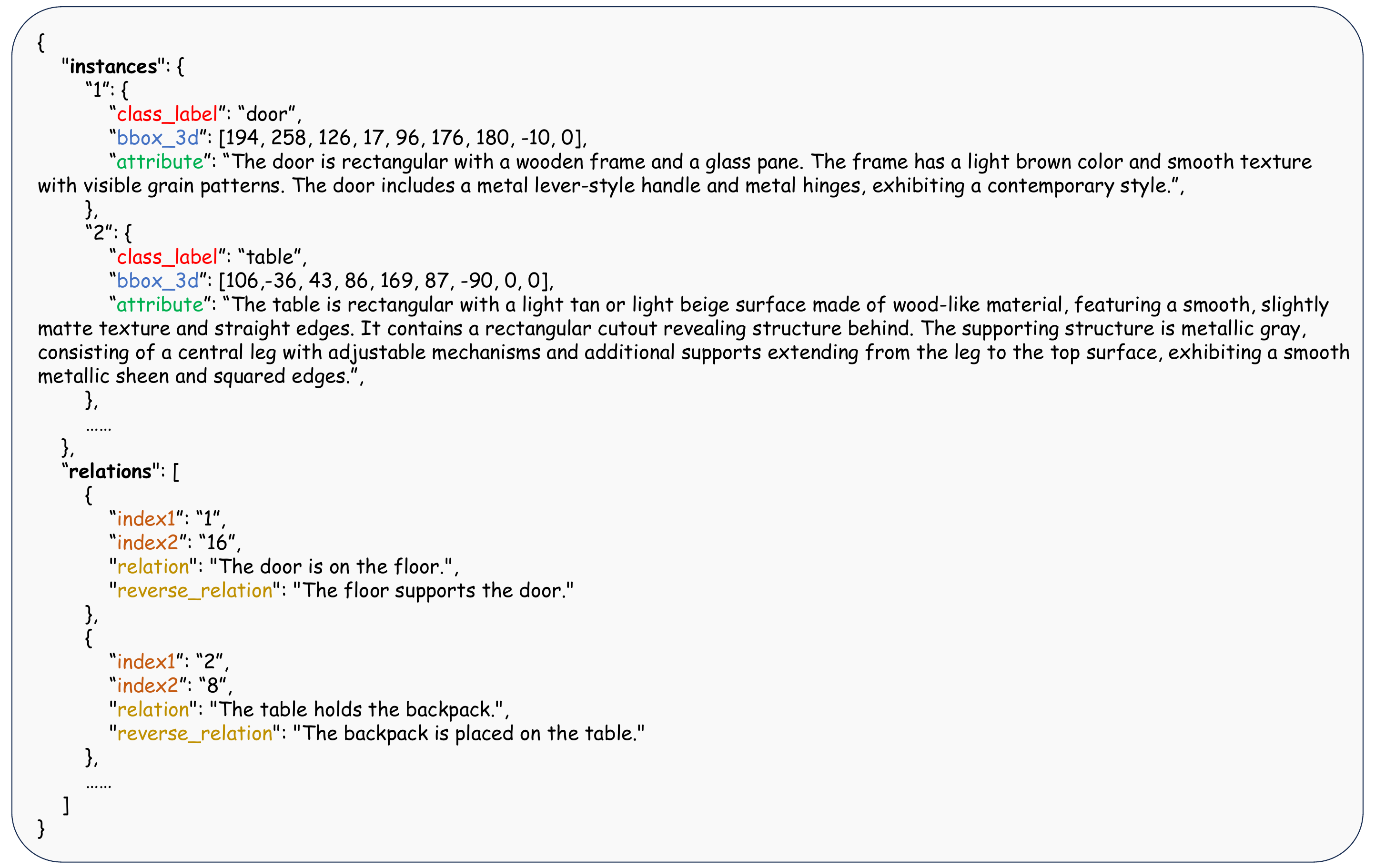}
\vskip -0.1in
\caption{
An example of holo-captioning data in JSON format. 
A holo-caption is a structured textual description that comprehensively depicts all entities within a 3D scene, including their semantic tags, spatial locations, attributes, and inter-entity relations.
Best viewed in color.
}
\label{fig:taskexample}
\vskip -0.1in
\end{figure}

In this part, we provide an example to clearly illustrate the task formulation of our proposed holo-captioning in greater detail. 
Since holo-captions are structured textual descriptions, they can be naturally expressed in JSON format for clarity and parsability, as illustrated by the example in Fig.~\ref{fig:taskexample}. 
Here, we present two entity instances within a scene, along with the relations between two specific pairs of instances. 
Because a full scene typically contains numerous entity instances and relation pairs, we omit the remainder for clarity and due to space constraints.
As shown in Fig.~\ref{fig:taskexample}, our holo-captioning includes a semantic tag, a precise 9-DoF bounding box, and a detailed attribute description for each entity instance in the 3D scene. 
Additionally, it includes a relation description for each instance pair that has a relation. 
Notably, our attribute descriptions encompass a wide range of attribute types, \eg, color, shape, material, texture, and constituent parts, making them much more comprehensive than those in previous 3D captioning benchmarks (\eg, Scan2Cap~\cite{Chen2021Scan2Cap}, which consists of only brief descriptions like ``the table is black'').

\subsection{Properties of our Holo-captioning Formulation}

In this part, we discuss the properties of our holo-captioning formulation, which serves as the initial effort toward our concept of ``text equivalence''. 
At its core, our formulation decomposes 3D scenes into four fundamental aspects, namely semantic tags, spatial locations (parameterized as 3D bounding boxes), attribute descriptions, and inter-entity relation descriptions. 
This design ensures a highly comprehensive representation of a 3D scene, and also maintains efficiency by strategically omitting inferable details. 
To demonstrate the comprehensiveness of our formulation, we provide three representative examples illustrating how the strategically omitted details can be accurately derived from the provided information. 
\begin{itemize}
  \item First, holo-captions omit location descriptions in pure words (\eg, ``A is at the center of the scene'').
Instead, we employ 9-DoF bounding boxes to describe entity locations. 
These bounding boxes are much more precise than vague textual descriptors, and allow us to derive location descriptions in natural language. 
  \item Second, we omit distance descriptions between entity instances (\eg, ``A is 5 meters away from B''), since this information can be computed from instance boxes, and including it would introduce redundancy into holo-captions. 
  \item Third, we exclude position-aware relational descriptions (\eg, ``A is on the right corner of B''), retaining only position-agnostic relational descriptions (\eg, ``A is on B''). 
This design choice is motivated by two factors: 
(1) 3D scenes are rotation-invariant, making terms like ``left'' or ``right'' inherently ambiguous; 
and (2) fine-grained, position-aware relations can be deterministically inferred by combining the exact 3D bounding boxes with basic position-agnostic relations. 
\end{itemize}

\section{More Details of HoloScan}
\label{sec:holoscandetails}
\vskip -0.05in

\subsection{Human Curation}
To construct a high-quality test set for evaluating holo-captioning models, we design a rigorous human-in-the-loop annotation refinement procedure to rectify the semantic labels, attribute descriptions, and inter-entity relation descriptions produced by (M)LLMs.
We keep the 3D bounding boxes fixed as spatial anchors, as these bounding boxes have been previously human-annotated in established benchmarks and datasets. 
The procedure consists of three primary stages supported by a custom Open3D-based visualization tool that renders point clouds and projects multi-view 2D RGB observations. 
First, by utilizing the 3D bounding boxes, we locate the corresponding entity instances within the 3D scene and incorporate multi-view images to verify whether the semantic tags are correctly annotated.
Any identified errors are subsequently corrected by human experts.
In the second step, with both the multi-view images and 3D scene content as references, we validate whether the attribute descriptions are correct.
This involves verifying not only the accuracy of the existing content but also the comprehensiveness of the descriptions (\ie, whether any attributes have been missed). 
Concurrently, we strictly regulate the terminology used in the descriptions.
For example, view-dependent terms such as ``left'' and ``right'' should be avoided, because they are not invariant under scene rotation. 
Finally, based on the same 2D and 3D context, we refine pairwise relational descriptions.
Specifically, these relational annotations strictly prioritize physical-contact relations.
If no explicit interaction is observed, the description is simplified to a standard adjacency state (\eg, ``A is adjacent to B''), and references to unassociated third-party objects are strictly prohibited to maintain purely pairwise contextualization. 
Additionally, we enforce semantic consistency between a relation description and its reverse (\ie, the (A, B) and (B, A) counterparts). 
Overall, this rigorous annotation refinement procedure yields a reliable test set for model evaluation.

\subsection{Reliability} 

The reliability of our HoloScan benchmark rests on two key factors, namely the effectiveness of our proposed data engine, HoloEngine, and the rigorous human curation procedure applied to the test set.
First, our proposed HoloEngine is specifically tailored to our proposed holo-captioning task. 
To effectively handle scene complexity, we propose to produce captions in an instance-centric manner, decomposing holistic, complex scenes into discrete, manageable elements for annotation. 
This strategy significantly enhances annotation quality, as generating accurate captions for isolated entity instances is demonstrably easier for (M)LLMs than parsing and comprehensively describing highly cluttered scenes in a single pass. 
Additionally, we systematically select the most capable (M)LLM for each specific stage of HoloEngine, considering the unique  characteristics of the data and each stage. 
We validated these selections through pilot experiments, wherein human experts meticulously evaluated the generation quality of various (M)LLMs on a compact yet diverse subset to identify the most suitable candidate for each subtask.
Regarding the second factor, namely the reliability of the test set, we have already elaborated on the human curation procedure in the preceding subsection, and thus we omit the details here.
Crucially, the high quality of our data production pipeline is further verified by compelling empirical evidence. 
Notably, we observe a strong alignment in the performance trends of various models when evaluated on the auto-generated validation set versus the human-curated test set. 
This consistency serves as strong evidence of the high quality of our generated annotations. 
More importantly, models trained on the data produced by HoloEngine achieve substantial performance gains over previous state-of-the-art methods. 
This superiority validates the effectiveness of our HoloEngine and the reliability of the HoloScan benchmark.

\section{More Implementation Details}
\label{sec:moreimplementation}
\vskip -0.05in

\subsection{HoloScribe Details}
Our proposed HoloScribe model adopts an ``Encoder-MLP-LLM'' architecture following SpatialLM~\cite{mao2025spatiallmtraininglargelanguage}. 
Specifically, HoloScribe adopts Sonata~\cite{DBLP:conf/cvpr/0002DFSXYENZS25} as the vision encoder, Qwen2.5-0.5B~\cite{qwen2025qwen25technicalreport} as the LLM, and a two-layer MLP as the projector. 
The model is initialized from the SpatialLM1.1-Qwen-0.5B checkpoint, and finetuned on the training set of HoloScan for one epoch using LoRA (rank $r=64$ and $\alpha=128$). 
We use the AdamW optimizer with a batch size of 12 and a learning rate of $5\times 10^{-4}$.
Since HoloScribe decomposes scene captions into multiple aspects, we use three types of QA-formatted point-text data for training, which are detailed below. 
We mix these three types of point-text data together for training models, and the training loss is the standard auto-regressive loss (\ie, next token prediction). 
During inference, we generate holo-captions step-by-step using the same prompts as in training.

\noindent \textbf{Grounded Instance Discovery.}
For each 3D scene, we construct point-text tuples in the format of $(\mathbf{Q}_v,\mathbf{Q}_{\text{inst}},\mathbf{H}_{\text{inst}})$ for training, where $\mathbf{Q}_v$ is the input 3D scene in point cloud format, $\mathbf{Q}_{\text{inst}}$ denotes the prompt and $\mathbf{H}_{\text{inst}}$ is a holo-instance text composed of individual instance texts. 
Given that real-world scenes often contain dozens of instances, we divide these instances into three types, namely \textit{architectural elements}, \textit{common objects}, and \textit{other objects}, to improve model prediction.  
This results in several point-text tuples per scene, and we use different prompts to instruct HoloScribe to generate the desired output. 
These prompts are:
\begin{tcolorbox}[
  colback=gray!10,    
  colframe=black,      
  boxrule=0.2pt,       
  arc=2pt,             
  left=4pt,right=4pt,top=4pt,bottom=4pt,
  fontupper=\comicneue\fontsize{7.5pt}{9pt}\selectfont
]
\begin{itemize}
    \item \textbf{Detect wall, door and window instances.}
    \item \textbf{Detect floor and ceiling instances.}
    \item \textbf{Detect basic instances.}
    \item \textbf{Detect more instances except basic categories.}
\end{itemize}
\end{tcolorbox}
\noindent The first two prompts are for architectural elements, the third covers common objects, and the fourth handles other objects.
We treat the categories recognizable by SpatialLM~\cite{mao2025spatiallmtraininglargelanguage} as common objects.

\noindent \textbf{Anchor-aware Instance Linking.}
This phase yields a set of point-text tuples $\{(\mathbf{Q}_v,\mathbf{Q}_\text{pair},\mathbf{H}_{\text{subg}},\mathbf{H}_{\text{pair}})\}$ for training, where $\mathbf{Q}_\text{pair}$ is a tailored prompt, and $\mathbf{H}_\text{subg}$ is text describing a sampled subgraph. 
$\mathbf{H}_\text{pair}$ is a relation text specifying relationship existence, serving as the target output in this phase.  
The prompt is given as follows:
\begin{tcolorbox}[
  colback=gray!10,   
  colframe=black,      
  boxrule=0.2pt,       
  arc=2pt,             
  left=4pt,right=4pt,top=4pt,bottom=4pt, 
  fontupper=\comicneue\fontsize{7.5pt}{9pt}\selectfont
]
\begin{itemize}
    \item \textbf{Predict relation pairs among boxes as follows: \{\}.}
\end{itemize}
\end{tcolorbox}
\noindent In this prompt, the placeholder ``\texttt{\small{\{\}}}'' is filled with the instance text of the sampled subgraph.

\noindent \textbf{Grounded Description Generation.}
For the attribute dimension, we construct point-text tuples $\{(\mathbf{Q}_v,\mathbf{Q}_\text{attr},\mathbf{H}_i,\mathbf{H}^a_i)\}$ for training, where $\mathbf{Q}_\text{attr}$ is the prompt. 
$\mathbf{H}_i$ is a text indicating the semantic tag and spatial location of the $i$-th ground-truth instance, resembling holo-instance texts. 
$\mathbf{H}^a_{i}$ is the corresponding ground-truth attribute description, which serves as the target output of the attribute dimension. 
The prompt is given as follows:
\begin{tcolorbox}[
  colback=gray!10,    
  colframe=black,       
  boxrule=0.2pt,    
  arc=2pt, 
  left=4pt,right=4pt,top=4pt,bottom=4pt, 
  fontupper=\comicneue\fontsize{7.5pt}{9pt}\selectfont
]
\begin{itemize}
    \item \textbf{Comprehensively describe the attributes of the instance as follows: \{\}. Please output the description:}
\end{itemize}  
\end{tcolorbox}
\noindent In this prompt, the placeholder ``\texttt{\small{\{\}}}'' is filled with the instance text of the specified instance. 
Additionally, this phase constructs point-text tuples $\{(\mathbf{Q}_v,\mathbf{Q}_\text{rel},\mathbf{H}_{i,j},\mathbf{H}^r_{i,j})\}$ for training relation description prediction, where $\mathbf{Q}_\text{rel}$ is the prompt. 
$\mathbf{H}_{i,j}$ is the instance text composed of the $i$-th and $j$-th instances, and $\mathbf{H}^r_{i,j}$ is the corresponding ground-truth relation description. 
The prompt is given as follows:
\begin{tcolorbox}[
  colback=gray!10,   
  colframe=black,      
  boxrule=0.2pt,       
  arc=2pt,             
  left=4pt,right=4pt,top=4pt,bottom=4pt, 
  fontupper=\comicneue\fontsize{7.5pt}{9pt}\selectfont
]
\begin{itemize}
    \item \textbf{Describe the relation between the two boxes as follows: \{\}}
\end{itemize}
\end{tcolorbox}
\noindent In this prompt, the placeholder ``\texttt{\small{\{\}}}'' is filled with the instance text of the specified pair of instances.

\subsection{Baseline Details}
In our main experiments, we use three types of models for comparison, namely 3D dense captioners, 3D LLM generalists, and holo-captioning specialists. 
For 3D dense captioners (\eg, Vote2Cap-DETR~\cite{Chen2023Vote2Cap}) and 3D LLM generalists (\eg, LL3DA~\cite{Chen2024CVPR}), we directly use the released checkpoints and apply them to holo-captioning following their default dense captioning pipelines. 
To establish holo-captioning specialists (\eg, SpatialLM-Tuned), we use the training data from HoloScan to finetune 3D LLMs (\eg, SpatialLM~\cite{mao2025spatiallmtraininglargelanguage}), thereby endowing them with holo-captioning capabilities. 
More specifically, to improve holo-captioning performance, we adopt a simple decoupled pipeline by performing attribute and relation prediction separately, as holo-captions for 3D scenes are usually long and information-dense. 
As shown in Table~3 in the main manuscript, our HoloScribe significantly outperforms these baselines, which is attributed to our proposed instance-aware decoupled pipeline.

\section{More Experiment Results}
\vskip -0.05in

\subsection{Scene Reconstruction}
\label{sec:morereconstruction}

\begin{figure}[t]
\centering
\includegraphics[width=0.95\linewidth]{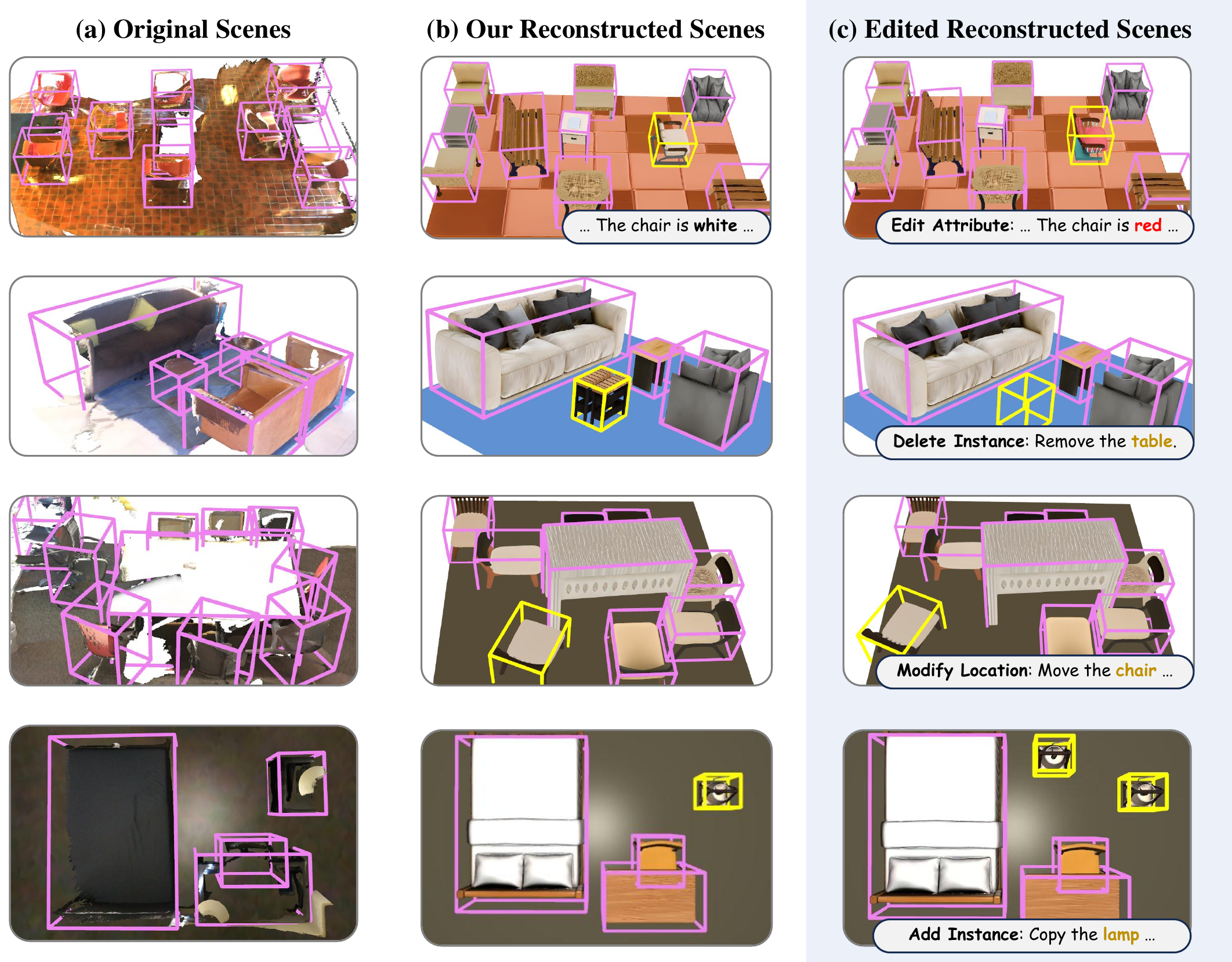}
\vskip -0.1in
\caption{
Qualitative analysis based on scene reconstruction.
We show several examples of real scenes from HoloScan, along with their text-based reconstructed scenes and the corresponding edited reconstructed scenes. 
Entity instances are highlighted with \underline{\textcolor[RGB]{238,130,238}{violet}} boxes, while \underline{\textcolor[RGB]{255,200,0}{yellow}} boxes mark the instances selected for editing in the reconstructed scenes and their edited counterparts in the edited scenes.
Best viewed in color.
}
\label{fig:reconstructionsupp}
\vskip -0.1in
\end{figure}

In this part, we extend the experiments in the main manuscript by showcasing more scene reconstruction results, which qualitatively demonstrate the effectiveness and utility of our proposed HoloScribe model. 
Specifically, given an input 3D scene, we first use HoloScribe to generate a holo-caption, which serves as a textual representation of the scene. 
Next, we employ Hunyuan3D~\cite{DBLP:journals/corr/abs-2506-15442} to generate 3D assets based on the semantic tags and attributes of entity instances extracted from the holo-caption.
We further refine the size and pose of each instance using the predicted bounding boxes and place them in 3D space to assemble the reconstructed scene.
To enhance visualization quality, all floor instances (except the first example) are rendered using solid-colored cubes. 
As shown in Fig.~\ref{fig:reconstructionsupp}~(b), HoloScribe produces reconstructed scenes that resemble the original scenes, particularly in terms of accurate semantic tagging and spatial localization. 

Additionally, we conduct a scene editing experiment to showcase a practical application enabled by our holo-captioning.
Specifically, to edit a reconstructed scene, we simply modify the generated holo-caption and then reconstruct a new scene based on the edited text. 
Fig.~\ref{fig:reconstructionsupp}~(c) presents four scene editing examples, namely attribute editing, instance deletion, instance relocation, and instance addition. 
As shown in the figure, our holo-captioning enables flexible, instance-level modifications while keeping all non-edited instances identical to those in the unedited scene.
Together, these experiments demonstrate the versatility and practicality of our holo-captions.

\subsection{Human Consistency Analysis for HoloScore}
Following Dong et al.~\cite{DBLP:journals/corr/abs-2405-19092}, we conduct a consistency analysis between human ratings and HoloScore values using Pearson correlation (PCC), Kendall's $\tau$, and sample-wise $\tau$.
As shown in Table~\ref{tab:human-consistency}, HoloScore aligns much better with human judgments than BLEU/CIDEr-based metrics.
\begin{table}[h]
\vskip -0.1in
\fontsize{8.5pt}{11.5pt}\selectfont
\setlength\tabcolsep{3.0pt}
\caption{
Human consistency analysis for HoloScore.
We report Pearson correlation (PCC), Kendall's $\tau$, and sample-wise $\tau$ between metric scores and human ratings.
\textbf{Bold} highlights the best results.
}
\vskip -0.1in
\label{tab:human-consistency}
\centering
\begin{tabular}{c | c c c}
\hline
Metrics & ~~PCC ($\rho$) $\uparrow$~~ & ~~K-$\tau$ $\uparrow$~~ & ~~Sw-$\tau$ $\uparrow$~~ \\
\hline
~~~BLEU~\cite{DBLP:conf/acl/PapineniRWZ02} + 3D Box IoU~~~  
& ~0.35~ & ~0.33~ & ~0.38~ \\
CIDEr~\cite{DBLP:conf/cvpr/VedantamZP15} + 3D Box IoU 
& ~0.39~ & ~0.35~ & ~0.40~ \\
\rowcolor{pink!20}
HoloScore & \textbf{0.60} & \textbf{0.55} & \textbf{0.56} \\
\hline
\end{tabular}
\vskip -0.1in
\end{table}
Additionally, the default thresholds of HoloScore are set based on preliminary human-consistency experiments.
While threshold changes affect absolute metric scores (\eg, increasing all thresholds by $+0.05$ causes a $9\%$ overall drop), the relative rankings of all models remain unchanged across multiple tested thresholds, demonstrating the robustness of our metric.
Equal weighting is adopted as a parameter-free default for simplicity.
Our experimental conclusions do not rely solely on this aggregation, as our model consistently performs best on each sub-metric.
In practice, users can adjust these weights as needed.

\subsection{Downstream Robotics Applications}

\begin{figure}[t]
\centering
\includegraphics[width=0.9\linewidth]{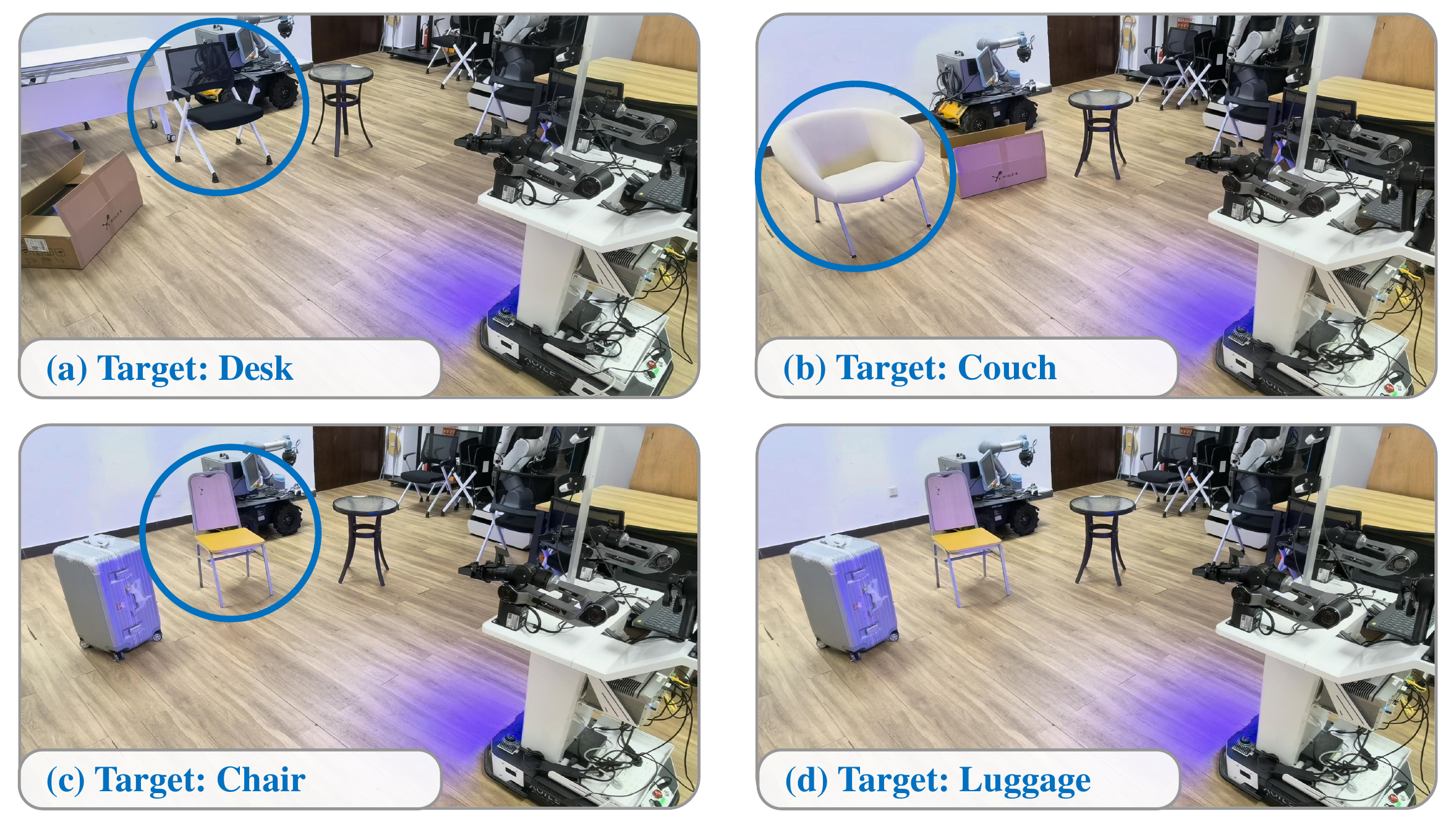}
\vskip -0.1in
\caption{
Demonstration of four real-world scenes for open-vocabulary object navigation.
Best viewed in color.}
\label{fig:objectnav}
\vskip -0.1in
\end{figure}

To demonstrate the utility of our HoloScribe, we deploy it in two real-world downstream robotics tasks, namely open-vocabulary object navigation~\cite{DBLP:journals/corr/abs-2008-09403} and mobile manipulation~\cite{10.1115/1.4054611}. 
In these downstream tasks, our HoloScribe is employed to generate holo-captions for scenes, from which we extract entity detection results and use them for performing robotics tasks. 
The hardware platform utilizes an Agilex Cobot Magic robot, equipped with two Intel RealSense D435 RGB-D cameras. 
These two cameras are referred to as the global and local cameras, serving the navigation and manipulation tasks, respectively. 

In the open-vocabulary object navigation task, the robot plans and navigates to the nearest instance of a specified object class. 
First, the scene point cloud is acquired via the global RGB-D camera and transformed from the camera coordinate system to the robot's local frame using a hand-eye transformation matrix. 
Then, the aligned scene point cloud is fed into our HoloScribe model to generate holo-captions, from which the entity detection results are extracted. 
Given a specific object label in text format, the system first retrieves the detected entity instances that match the label and then identifies the target instance based on Euclidean distance. 
If multiple instances match the given label, the closest one is selected as the target.
Based on the center coordinates of the identified target instance, the system further calculates the positional deviation ($\Delta x$, $\Delta y$) and the angular deviation ($\Delta \theta$) between the current robot position and the target position. 
These deviations are then used to generate a trajectory to drive the robot toward the target, thereby accomplishing the object navigation task. 
In this experiment, we evaluate the open-vocabulary object navigation task across four scenes, each featuring a distinct target object, as shown in Fig.~\ref{fig:objectnav}. 
For each scene, we command the robot to execute the task 10 times and report the average success rate.
Task execution strongly depends on the entity instances detected by HoloScribe.
Table~\ref{tab:objectnav-comparison} reports the performance comparison in terms of average success rate.
As shown in the table, HoloScribe achieves a 75\% overall average success rate, significantly surpassing the SpatialLM baseline (25\%). 
This demonstrates the superiority of our proposed HoloScribe model. 

\begin{table}[h]
\vskip -0.15in
\fontsize{7.5pt}{10.5pt}\selectfont
\setlength\tabcolsep{3.0pt}
\centering
\caption{Comparison between our HoloScribe and SpatialLM in open-vocabulary object navigation. 
The performance is measured in terms of success rate (\%).}
\label{tab:objectnav-comparison}
\vskip -0.1in
\begin{tabular}{c|cccc|c} 
\hline
\textbf{Models} & Scene (a) & Scene (b) & Scene (c) & Scene (d) & Overall \\
\hline
SpatialLM~\cite{mao2025spatiallmtraininglargelanguage} 
& 100.0 & 0.0 & 0.0 & 0.0 & 25.0 \\
HoloScribe (Ours) 
& 100.0 & 100.0 & 100.0 & 0.0 & 75.0 \\
\hline
\end{tabular}
\vskip -0.15in
\end{table}

In the mobile manipulation task, the robot builds upon object navigation to transfer small objects between target locations within the scene. 
For the navigation component, the key difference is that the system initially determines both the starting and terminal positions.
During the manipulation phase, we deploy a framework inspired by CaP~\cite{DBLP:journals/corr/abs-2209-07753}. 
Given visual prompts captured by the local camera, we employ an MLLM (\ie, Qwen3-VL-Plus~\cite{DBLP:journals/corr/abs-2511-21631}) to accomplish two objectives: selecting an appropriate action primitive (\eg, grasping, placing) from a predefined library, and predicting interaction keypoints as 2D pixel coordinates. 
These 2D points are subsequently projected into 3D space to determine the target spatial positions for the robotic arm. By composing the selected primitive with the corresponding 3D keypoints, the system seamlessly executes complex manipulation tasks.
In the project page, we provide a demonstration video of the mobile manipulation process. 
This demonstration further validates the robustness of HoloScribe in real-world scenarios.

\begin{table}[b]
\fontsize{7.5pt}{10.5pt}\selectfont
\setlength\tabcolsep{3.0pt}
\caption{
Quantitative analysis of our HoloScribe based on different LLMs. 
We report the results on the validation set of HoloScan. 
Performance is measured by HoloScore, and we report the scores in tagging, location, attribute, relation, and the overall score. 
\textbf{Bold} highlights our results. 
}
\vskip -0.1in
\label{tab:sota-llama}
\centering
\begin{tabular}{c | c | c  c  c c | c }
\hline
Models & LLM & Tagging & Location & Attribute & Relation & Overall \\
\hline
SpatialLM-Tuned~\cite{DBLP:conf/icml/HuangYMLLW0ZJ024} & Qwen2.5-0.5B
& 51.73 & 9.94  & 6.74  & 2.60 & 71.01 \\
LL3DA-Tuned~\cite{Chen2024CVPR} & Qwen2.5-0.5B
& 31.25 & 16.50 & 10.36 & 4.83 & 62.94 \\
\rowcolor{pink!20}
HoloScribe (Ours) & ~Qwen2.5-0.5B~
& \textbf{60.43} & \textbf{22.33} & \textbf{22.52} & \textbf{11.08} & \textbf{116.36} \\
\rowcolor{pink!20}
HoloScribe (Ours) & ~Llama3.2-1B~
& \textbf{60.24} & \textbf{25.75} & \textbf{23.47} & \textbf{9.51}  & \textbf{118.97} \\
\hline
\end{tabular}
\vskip -0.15in
\end{table}

\subsection{Quantitative analysis with different LLMs}
In this part, we conduct a quantitative analysis to verify the generalizability of our HoloScribe method across different LLMs. 
While our primary experiments utilized Qwen2.5-0.5B, here we extend our analysis to Llama3.2-1B.
Both backbones were initialized from the SpatialLM~\cite{mao2025spatiallmtraininglargelanguage} checkpoints (\ie, SpatialLM1.1-Qwen-0.5B and SpatialLM1.1-Llama-1B), and then finetuned on the training set of HoloScan. 
As shown in Table~\ref{tab:sota-llama}, our model with Llama3.2 achieves performance comparable to that with Qwen2.5, with both configurations significantly outperforming previous state-of-the-art models. 
These results demonstrate that the effectiveness of HoloScribe is not dependent on a specific backbone, thereby confirming its robustness.

\begin{figure}[t]
\vskip -0.1in
\centering
\includegraphics[width=1\linewidth]{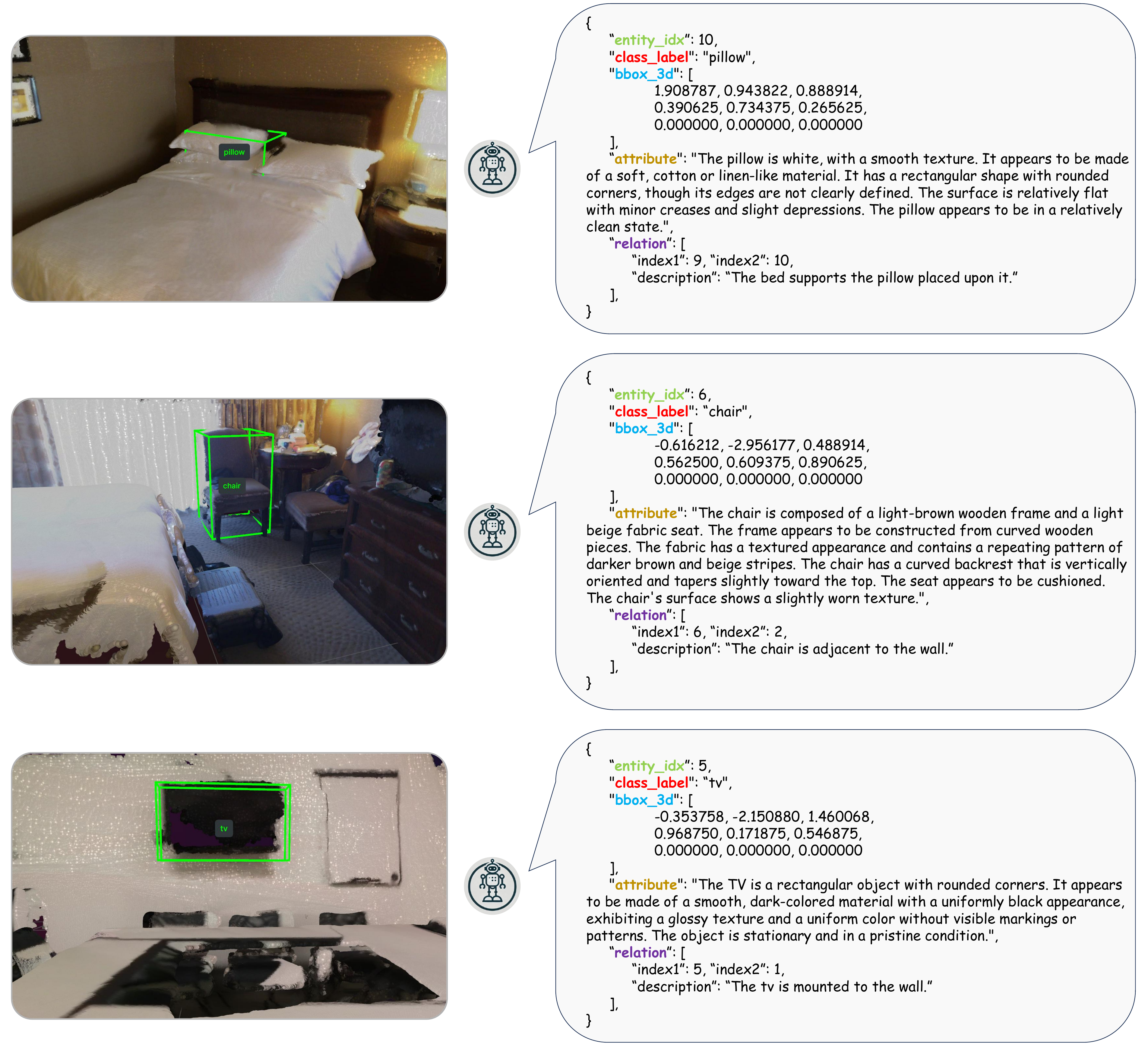}
\vskip -0.1in
\caption{
Example outputs of our HoloScribe model in JSON format. 
For each sample, we visualize the entity instance by rendering the raw point cloud of the 3D scene, with a 3D bounding box employed to localize the target instance.
Additionally, we show the text from the 3D scene's predicted holo-caption that is relevant to the specified target instance. 
Note that due to space constraints, presenting the full holo-caption of a scene in the manuscript is impractical given the large number of entity instances involved, thus we only showcase one representative example for illustration.
Best viewed in color.
}
\label{fig:modelexample}
\vskip -0.1in
\end{figure}

\subsection{Qualitative Examples}
In this part, we show qualitative examples to demonstrate the holo-captioning capabilities of our HoloScribe model. 
Fig.~\ref{fig:modelexample} presents three examples of holo-captions, where each example corresponds to a specific entity instance in a 3D scene. 
As shown in the figure, our model generates holo-captions that are largely accurate and satisfactory for entity instances.

\bibliographystyle{splncs04}
\bibliography{main}

\end{document}